\documentclass[10pt]{article}

\usepackage{amsmath}
\usepackage{graphicx}
\usepackage{epstopdf}
\usepackage{comment}
\usepackage{hyperref}
\usepackage[margin=0.8in]{geometry}

\begin{document}

\title{Photorealistic Video Style Transfer}
\author{Report by Michael Honke, Rahul Iyer and Dishant Mittal}
\maketitle

\section{Introduction}

It entails huge amount of time for a professional artist to re-draw an image manually in a particular artistic style. Far more difficult than this is to accomplish this task for a certain video sequence. Nowadays, the advancement in computing technology has revolutionalized many aspects of problem solving. In this project we show how to transfer the style from one photorealistic image to a whole video sequence using deep learning. We make use of two of the most prominent works in style transfer which are as follows:

\begin{itemize}
\item Ruder et al. [1] - They portrayed a technique which enables to transfer the style from an artistic image (for instance, a painting) to a complete video sequence. Basically, they extended the previous research by 
Gatys et al. [2] and Johnson et al. [3] to complete video sequences. Their technique transferred the style captured from an artistic image to the entire video. They deduced that independently processing each frame of the video leads to glimmering and untrue irregularities. This was because the solution of the style transfer task was not steady. With the goal of stabilizing the transfer process
and preserving an even changeover between independent video frames, they established a temporal constraint
that penalized divergence between two consecutive
frames. The temporal constraint introduced by them considered the optical flow present in the original video and in place of inflicting a penalty to the deviations from the previous frame, they applied penalty to the 
divergence along the trajectories of points. The regions
which were not concealed and the boundaries of the motion were eliminated from the penalizer. This imparted the liberty to rerender the unconcealed regions and deformed motion boundaries while conserving the look of the rest of the image.

\item Luan et al. [4] - They portrayed a deep-learning technique particularly for style transfer in photorealistic images that was able to control a huge diversity of image content while reliably transplanting the style from reference image. Their cardinal contribution was to impel the process of conversion from the input to the output to be locally affine in colorspace, and to depict this restriction as a custom and completely differentiable loss function (which they call as energy term). They proved that their technique successfully subdues the deformations and produces convincing photorealistic style transfers in a wide variety of outlines, including transfer of the parameters like time of day, season, weather and artistic edits. They utilised the Matting Laplacian to restrict the conversion from the input to the output to be locally
affine in colorspace. Furthermore, they used semantic segmentation which instigated more meaningful style transfer.
\end{itemize}

In the implementation of this project we merge the loss functions used by Luan et al. [4] and Ruder et al. [1] to get a cumulative loss function used for optimizing the style transimmision from photorealistic image to video.

\section{Loss Functions}

\begin{itemize}
\item Ruder et al. [1] - The main aim is to render a stylized image x describing the style of an image a and the content of an image p.
Gatys et al. [2] devised an energy minimization problem
consisting of a content loss and a style loss. The
central concept is that attributes extracted by a convolutional neural network contain details about the content of the image,
whereas the correlations of these attributes store the details related to style. Johnson et al. [3] depicted that in lieu of solving an optimization problem, a faster technique would be to straightaway learn a function for style transfer for one specific style mapping through an input image
to its corresponding stylized image. Convolutional Neural Network could be used to express such a function using the parameters w. The
expected loss can be minimized by tranining the network over a random image. Ruder et al. proposed two losses by considering the approaches proposed by Gatys and Johnson:
\begin{itemize}

\item Short term loss:  \[ Loss_{shortterm}(f^{(i)} ,a,x^{(i)}) = \alpha Loss_{content}(f^{(i)},x^{(i)}) \]

\[+ \beta Loss_{style}(a,x^{(i)}) + \gamma Loss_{temporal}(x^{(i)}, w_{i-1}^{i}(x^{i-1}),c^{i-1,i}) \]

where i denotes index of a frame, f\textsuperscript{(i)} is the i\textsuperscript{th} frame of the video, a is the style image, x\textsuperscript{(i)} is the i\textsuperscript{th} stylized frame to be generated , c denotes weight, w is the function that warps a given frame using the optical flow field that was estimated between two images.

\item Long term loss:
\[ Loss_{shortterm}(f^{(i)} ,a,x^{(i)}) = \alpha Loss_{content}(f^{(i)},x^{(i)}) \]

\[+ \beta Loss_{style}(a,x^{(i)}) + \gamma \sum_{j \epsilon J, (i-j) \geq 1} Loss_{temporal}(x^{(i)}, w_{i-j}^{i}(x^{i-j}),c_{long}^{(i-j, i)}) \]

all other notations are same as defined previously, J denote the set of indices each frame should take into account relative to the frame number, e.g., with J = {1, 2, 4}, frame i takes frames i$-$1, i$-$2, and i$-$4
into account. 
\end{itemize}

In both of the above short and long term losses, L\textsubscript{content} and L\textsubscript{style} are defined by Gatys et al. [2]

\item Luan et al. [4] - Apart from the L\textsubscript{content} and L\textsubscript{style} described above, they explained how to
regularize these losses to preserve the structure of the input image and generate outputs that are photorealistic. Rather than directly imposing this constraint on the output image they applied it on the transformation which has been applied to the input image. Describing the space of photorealistic images is a probelm that remains unsolved. However, the authors proposed that we don't actually require to solve it if we leverage the fact that the input that would be fed into the model is already photorealistic. Their plan was to assure that this fact should not get lost during the transfer process by attaching a term to the equation of original loss function which penalizes image deformations. The proposed solution was to search for the transform of an image that is locally affine in color space. That means to search a function such that for every output spot there exists an affine function that maps the corresponding RGB components in input to their output equivalents. They build upon the Matting Laplacian of Levin et al. [5]. They came up with the following Loss function:
\[ Loss_m = \sum_{c = 1}^3 V_c [O]^T M_I V_c [O] \]

where M\textsubscript{i} is a standard linear system that only depends on the input image I and V\textsubscript{c}[O] is the vectorized version (N × 1) of the output image O.

One limitation of the style loss presented by Gatys et al. [2] was that  the Gram matrix is computed over the entire image. A precise distribution of neural responses is completely encoded by Gram matrix as it computes its vector components uptill an isometry. However, this can cause “spillovers” as that limits its power to adapt to variations in semantic context. The solution to this problem was that keeping the set of labels constant (i.e. sky, buildings, water, etc.) we can render image segmentation masks for both the input as well as reference images. This solution was similar to Neural Doodle [6] and semantic segmentation method [7]. Hence, they included the masks to the input image as supplementary channels and by appending the segmentation channels they built up the neural style algorithm. Finally that style loss was updated as follows

\[ Loss_{s+}^l = \sum_{c = 1}^C \frac{1}{2N_{l,c}^2} \sum_{i,j} (G_{l,c}[O]-G_{l,c}[S])_{i,j}^2 \]

where C is the number of channels in the semantic segmentation
mask. Finally they formulated the photorealistic style
transfer loss function by combining all 3 components losses together as:

\[ Loss_{total} = \sum_{1=1}^L\alpha_l L_c^l + \tau \sum_{1=1}^L\beta_l L_{s+}^l + \lambda L_m \]

where L is the total number of convolutional layers and l indicates the l\textsubscript{th} convolutional layer of the deep neural
network, $\tau$ is a weight that controls the style loss. $\alpha$ and
$\beta$ are the weights to configure layer preferences. $\lambda$ is a
weight that controls the photorealism regularization.

\end{itemize}

\section{Merging and Challenges}

To implement the style transfer from one photorealistic image to a whole video sequence using deep learning we merged the concepts from both the works i.e from Ruder et al. [1] and Luan et al. [4]. Precisely, we accomplished this in two parts: 

\begin{itemize}
\item We integrate the the loss functions used in style transfer in photorealistic images [4] to the loss functions used in artistic style transfer in videos [1].

\item The semantic segmentation used by Luan et al. [4] was done manually. We use the semantic segmentation technique proposed by Zhou et al. [7]. This helped us to automate the task of segmentation which can then be used in computing the overall loss for the videos. 
\end{itemize}

\section{Results}

\section{References}
[1] Ruder, M., Dosovitskiy, A., \& Brox, T. (2016, September). Artistic style transfer for videos. In German Conference on Pattern Recognition (pp. 26-36). Springer International Publishing.
\newline\newline
[2] Gatys, L. A., Ecker, A. S., \& Bethge, M. (2016). Image style transfer using convolutional neural networks. In Proceedings of the IEEE Conference on Computer Vision and Pattern Recognition (pp. 2414-2423).
\newline\newline
[3] Johnson, J., Alahi, A., \& Fei-Fei, L. (2016, October). Perceptual losses for real-time style transfer and super-resolution. In European Conference on Computer Vision (pp. 694-711). Springer International Publishing.
\newline\newline
[4] Luan, F., Paris, S., Shechtman, E., \& Bala, K. (2017). Deep Photo Style Transfer. arXiv preprint arXiv:1703.07511.
\newline\newline
[5] A. Levin, D. Lischinski, \& Y. Weiss. A closed-form solution
to natural image matting. IEEE Transactions on Pattern
Analysis and Machine Intelligence, 30(2):228–242, 2008.
\newline\newline
[6] S. Bae, S. Paris, \& F. Durand. Two-scale tone management
for photographic look. In ACM Transactions on Graphics
(TOG), volume 25, pages 637–645. ACM, 2006
\newline\newline
[7] Zhou, B., Zhao, H., Puig, X., Fidler, S., Barriuso, A., \& Torralba, A. (2017). Scene parsing through ade20k dataset. In Proc. CVPR.

\end{document}